\documentclass[runningheads]{llncs}
\usepackage{graphicx}
\usepackage{comment}
\usepackage{amsmath,amssymb} 
\usepackage{color}
\usepackage{hyperref}
\usepackage[para]{footmisc}

\usepackage{gensymb}
\usepackage{multirow}
\usepackage{multicol}
\usepackage{subfig}
\usepackage{float}
\usepackage{enumitem}
\usepackage{xcolor, colortbl}

\DeclareMathOperator*{\argmin}{arg\,min}

\begin{document}
\pagestyle{headings}
\mainmatter
\def\ECCVSubNumber{3632}  

\title{APRICOT: A Dataset of Physical Adversarial Attacks on Object Detection} 

%
\author{A. Braunegg \and
Amartya Chakraborty \and
Michael Krumdick \and
Nicole Lape \and
\mbox{Sara Leary} \and
Keith Manville\thanks{Corresponding author} \and
Elizabeth Merkhofer \and
Laura Strickhart \and
Matthew Walmer}
\authorrunning{A. Braunegg et al.}
%
\institute{The MITRE Corporation\\
\email{\{abraunegg, achakraborty, mkrumdick, nflett, sleary, kmanville, emerkhofer, lstrickhart, mwalmer\}@mitre.org}}
\maketitle

\begin{abstract}
Physical adversarial attacks threaten to fool object detection systems, but reproducible research on the real-world effectiveness of physical patches and how to defend against them requires a publicly available benchmark dataset. 
We present \mbox{APRICOT}\footnote[2]{Project page and dataset available at \href{https://apricot.mitre.org}{apricot.mitre.org}}, a collection of over 1,000 annotated photographs of printed adversarial patches in public locations.
The patches target several object categories for three COCO-trained detection models, and the photos represent natural variation in position, distance, lighting conditions, and viewing angle.
Our analysis suggests that maintaining adversarial robustness in uncontrolled settings is highly challenging but that it is still possible to produce targeted detections under white-box and sometimes black-box settings.
We establish baselines for defending against adversarial patches via several methods, including using a detector supervised with synthetic data and using unsupervised methods such as kernel density estimation, Bayesian uncertainty, and reconstruction error.
Our results suggest that adversarial patches can be effectively flagged, both in a high-knowledge, attack-specific scenario and in an unsupervised setting where patches are detected as anomalies in natural images.
This dataset and the described experiments provide a benchmark for future research on the effectiveness of and defenses against physical adversarial objects in the wild.
\end{abstract}

\section{Introduction}

\begin{figure}
   \centering
   \includegraphics[width=1.0\linewidth]{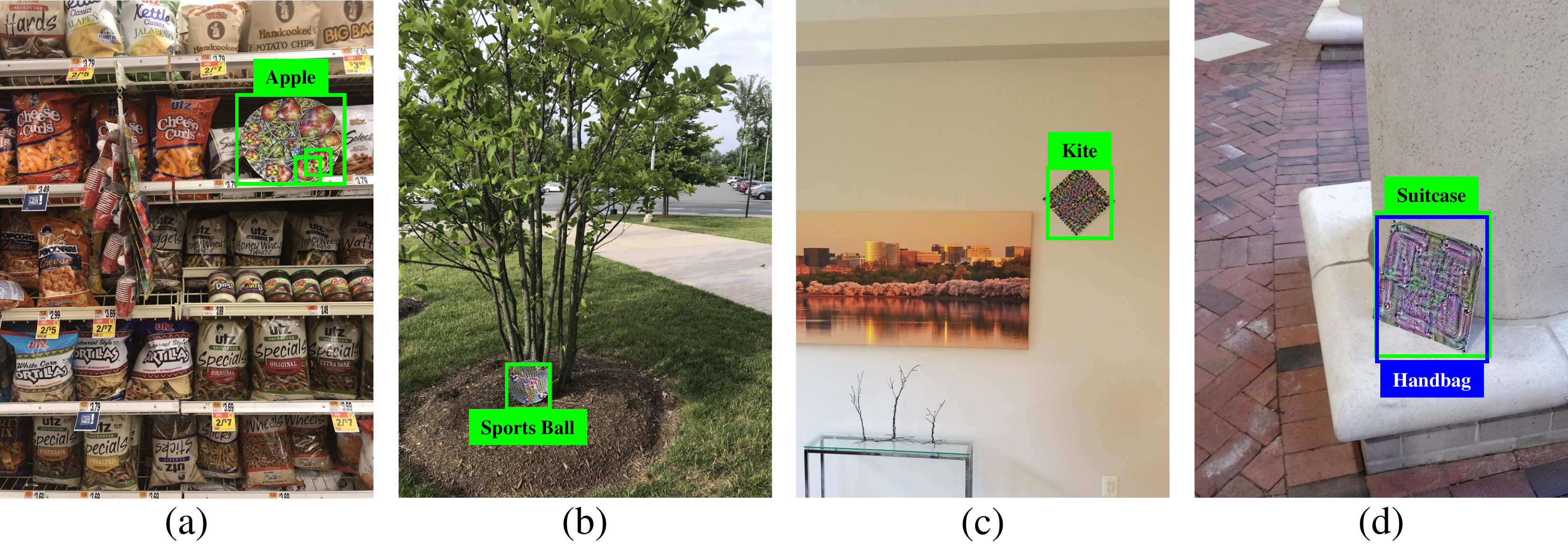}
   \caption{APRICOT images run through FRCNN; only detections caused by the patch are shown. (a) FRCNN adversarial patch targeting apple category. (b) SSD adversarial patch targeting person category. (c) RetinaNet adversarial patch targeting bottle category. (d) FRCNN adversarial patch targeting suitcase category.}
   \label{fig:patch_example}
\end{figure}

Image detection and classification models have driven important advances in safety-critical contexts such as medical imaging \cite{fu2019medical} and autonomous vehicle technology \cite{fridman2019vehicle}. However, deep models have been demonstrated to be vulnerable to adversarial attacks \cite{biggio2013evasion,szegedy2014intriguing}, images designed to fool models into making incorrect predictions. The majority of research on adversarial examples has focused on digital domain attacks, which directly manipulate the pixels of an image \cite{carlini2017towards,goodfellow2014explaining,moosavi2016deepfool,papernot2017practical}, but it has been shown that it is possible to make physical domain attacks in which an adversarial object is manufactured (usually printed) and then placed in a real-world scene. When photographed, these adversarial objects can trigger model errors in a variety of systems, including those for classification \cite{athalye2018synthesizing,brown2017adversarial,eykholt2018robust,kurakin2016adversarial}, facial recognition \cite{sharif2016accessorize}, and object detection \cite{chen2018robust,thys2019fooling}. These attacks are more flexible, as they do not require the attacker to have direct access to a system’s input pixels, and they present a clear, pressing danger for systems like autonomous vehicles, in which the consequences of incorrect judgements can be dire. In addition, most adversarial defense research has focused on the task of classification, while for real-world systems, defenses for deep detection networks are arguably more important.

Unlike digital attacks, physical adversarial objects must be designed to be robust against naturally occurring distortions like angle and illumination, such that they retain their adversariality after being printed and photographed under a variety of conditions. The key to achieving robustness is a method called Expectation over Transformation \cite{athalye2018synthesizing}, which directly incorporates simulated distortions during the creation of the adversarial pattern. Without this strategy, adversarial examples typically fail in the real world \cite{lu2017no}. For this reason, physical adversarial examples face a robustness-vs-blatancy trade-off \cite{chen2018robust}, as robust adversarial textures are typically more distinctive and visible to the human eye. This is in contrast to digital attacks, which only need to alter pixels by a tiny amount. It is a critical unanswered question whether making physical adversarial patches robust also makes them easy to detect using defensive systems.

Securing safety-critical detection systems against adversarial attacks is essential, but conducting research into how to detect and defeat these attacks is difficult without a benchmark dataset of adversarial attacks in real-world situations. We present APRICOT, a dataset of \underline{A}dversarial \underline{P}atches \underline{R}earranged \underline{I}n \underline{CO}n\underline{T}ext.  This dataset consists of over 1,000 images of printed adversarial patches photographed in real-world scenes. The patches in APRICOT are crafted to trigger false positive detections of 10 target COCO \cite{lin2014microsoft} classes in 3 different COCO-trained object detection models \cite{huang2017speed}. While prior works in physical adversarial examples have included some real-world experiments in fairly controlled environments \cite{brown2017adversarial,chen2018robust,eykholt2018robust,lu2017no}, our approach with APRICOT was to use a group of 20 volunteer photographers to capture images of adversarial patches in the wild. The dataset incorporates a wide variety of viewing angles, illuminations, distances-to-camera, and scene compositions, which all push the necessary level of patch robustness further than prior works. To the best of our knowledge, APRICOT is the first publicly released dataset of physical adversarial patches in context. APRICOT allows us to perform a comprehensive analysis of the effectiveness of adversarial patches in the wild. We analyze white-box effectiveness, black-box transferability \cite{papernot2017practical}, and the effect of factors such as patch size and viewing angle.

Despite the danger that physical adversarial examples pose against object detection networks, there has been limited research into defensive mechanisms against them. In addition to the APRICOT dataset, we present baseline experiments measuring the detectability of adversarial patches in real-world scenes. We adapt several digital domain defenses \cite{feinman2017detecting,metzen2017detecting,samangouei2018defense} to the task of defending a Faster-RCNN model \cite{ren2015faster} from black-box attacks in the real world. We consider two broad categories of defenses. ``Supervised defenses’’ like \cite{metzen2017detecting} assume that the defender is aware of the attacker’s method and can therefore generate their own attack samples and train a defensive mechanism with them. We show that, given this prior knowledge, it is possible to use images with synthetically inserted patches to train an adversarial patch detector with high precision. However, we believe that such defenses are insufficient in the long run, as they may not generalize as well to newly emerging attacks. ``Unsupervised defenses’’ like \cite{feinman2017detecting,samangouei2018defense} are attack-agnostic, operating without any prior knowledge of attack appearance. We test several unsupervised defense strategies, both for network defense and for adversarial patch localization. We hope that APRICOT will provide a testbed to drive future developments in defenses for object detection networks.

To summarize, our main contributions are as follows: (1) APRICOT, the first publicly available dataset of physical adversarial patches in context, (2) an assessment of the robustness of adversarial patches in the wild, as both white-box and black-box attacks, and (3) several supervised and unsupervised strategies for flagging adversarial patches in real-world scenes.
In Section \ref{sec:related}, we summarize related works in adversarial examples in both the digital and physical domains. In Section \ref{sec:dataset}, we describe the creation of the APRICOT dataset. In Section \ref{sec:eval}, we measure the effectiveness of the adversarial patches when placed in real-world scenes. Finally, in Section \ref{sec:detection}, we present several defenses designed to flag adversarial patches in the wild.

\section{Related Work} \label{sec:related}

\textbf{Adversarial Examples in Digital Images}
The phenomenon of adversarial examples in digital images has been explored extensively in recent years \cite{biggio2013evasion,carlini2017towards,goodfellow2014explaining,moosavi2016deepfool,szegedy2014intriguing}. Attack methods tend to follow a common framework of optimizing a perturbation vector that, when added to a digital image, causes the target classifier to produce an incorrect answer with high confidence. The most effective adversarial methods make perturbations that are imperceptible to humans. It is even possible to create ``black-box'' attacks without direct access to the target model by using a surrogate \cite{papernot2017practical}.

\textbf{Adversarial Examples in the Real World}
Digital domain attacks assume that the attacker is able to manipulate the pixels of an image directly. However, some works have expanded the scope of adversarial examples into the physical domain \cite{athalye2018synthesizing,brown2017adversarial,eykholt2018robust,kurakin2016adversarial,sharif2016accessorize}. Adversarial examples in the real world are particularly dangerous for applications that rely on object detection. \cite{thys2019fooling} developed a real-world adversarial patch that can block person detection. \cite{chen2018robust} developed an optimization framework called ShapeShifter that can create adversarial stop signs that will be detected but incorrectly classified. In creating the APRICOT dataset, we modify ShapeShifter to produce patches that cause detectors to hallucinate objects. Such patches could confuse machine learning systems and lead them to take pointless or dangerous actions that would be incomprehensible to a human observer.

\textbf{Defending Against Adversarial Examples}
Most prior adversarial defense literature focuses on digital attacks. In this work, we adapt several digital domain defenses with a focus on flagging physical adversarial examples. \cite{metzen2017detecting} proposed training a binary classifier to detect adversarial examples directly. This approach falls in the category of supervised defenses. They showed that training on one type of attack can transfer to weaker attacks, but not stronger ones. For unsupervised defense strategies, several authors have suggested that adversarial examples do not exist on the natural image manifold. \cite{feinman2017detecting} used kernel density estimation (KDE) to determine whether images are in low-density regions of the training distribution and Bayesian uncertain estimation with dropout to identify samples in low-confidence regions. PixelDefend\cite{song2018pixeldefend} uses a PixelCNN \cite{oord2016pixel} to model the natural image manifold and back-propagation to remove adversarial perturbations. Defense-GAN\cite{samangouei2018defense} uses a generative adversarial network (GAN)\cite{goodfellow2014generative} to model the image distribution and GAN inversion and reconstruction to remove perturbations. Real-world adversarial objects need to be much more blatant than digital attacks to be robust to natural distortions \cite{chen2018robust}. This suggests that physical attacks might be inherently more detectable, though this question has yet to be answered due to the lack of a dataset like APRICOT.

\section{The APRICOT Dataset} \label{sec:dataset}

APRICOT is a dataset of physical adversarial patches photographed in real-world scenes. APRICOT contains 1,011 images of 60 unique adversarial patches designed to attack 3 different COCO-trained detection networks. While previous works have tested physical adversarial objects under controlled settings, APRICOT provides data for testing them in diverse scenes captured by a variety of photographers. The dataset includes both indoor and outdoor scenes, taken at different times of day, featuring various objects in context, and with patch placements that vary in position, scale, rotation, and viewing angle.

The attacks in \cite{chen2018robust} and \cite{thys2019fooling} can be characterized as ``false negative attacks,” as they cause detectors to miss genuine objects. The attack style used in APRICOT is instead a ``false positive attack,” making detectors hallucinate objects that do not exist. Such attacks can be just as dangerous. For example, imagine that a malicious agent has designed a patch to trigger hallucination of the ``car” class and has placed it on the road. A self-driving car with a camera and detector model would see the patch, mistake it for a car, and suddenly stop or swerve to avoid a hallucinated collision. Furthermore, an untrained human observer would see the patch on the road but would not understand why the vehicle started to behave erratically. This would both threaten user safety and undercut user confidence.

The intended uses of APRICOT are twofold: firstly, to assess the risk posed by adversarial patches in real-world operating conditions, and secondly, to aid in the creation of new defenses.
Research is stalled by the lack of physical datasets, which are much more costly to create than digital datasets.
We contribute \mbox{APRICOT} as a benchmark for future works to use in developing and comparing defensive strategies for object detection networks.

\subsection{Generating Adversarial Patches}

To generate adversarial examples that are effective against detectors in the physical domain, we follow the approach from Chen et al. \cite{chen2018robust}, where the authors optimize physical adversarial examples constrained to the shape and color of a stop sign. Since one goal of our dataset is to promote understanding of real-world adversarial performance across a variety of target objects and arbitrary locations, we choose to drop the object-specific mask and color constraints. We also optimize them with random COCO images in the background. This makes our adversarial examples patches \cite{brown2017adversarial} that target detectors and are universal in that they can be placed in arbitrary locations in the real world.
The key algorithm behind these approaches is Expectation over Transformation \cite{athalye2018synthesizing}, which optimizes a given patch over many inputs, $X$, subject to a variety of transformations, $T$, making the adversarial nature of the patch robust over the range of transformations. Given a loss function $\mathcal{L}$ and detector $f$ and parameters $\theta$, we optimize to find a patch $\hat{p}$ with target category $y'$ according to:

\begin{align}
    \hat{p} = \argmin_{p \in {\rm I\!R}^{h \times w \times 3}}{\mathbb{E}_{x \sim X,t \sim T}[\mathcal{L} (f_\theta (t(x,p)),y')]}
    \label{eq:eot}
\end{align}

\begin{figure}[ht]
    \begin{center}
        \includegraphics[width=0.20\linewidth]{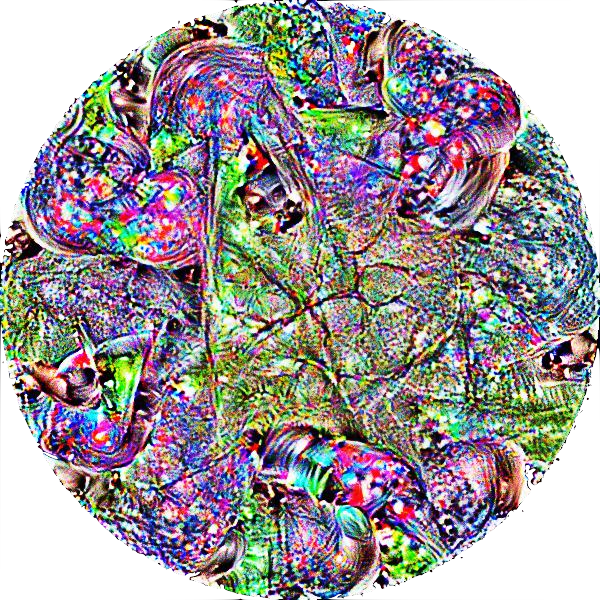}
        \includegraphics[width=0.20\linewidth]{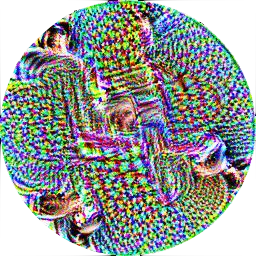}
        \includegraphics[width=0.20\linewidth]{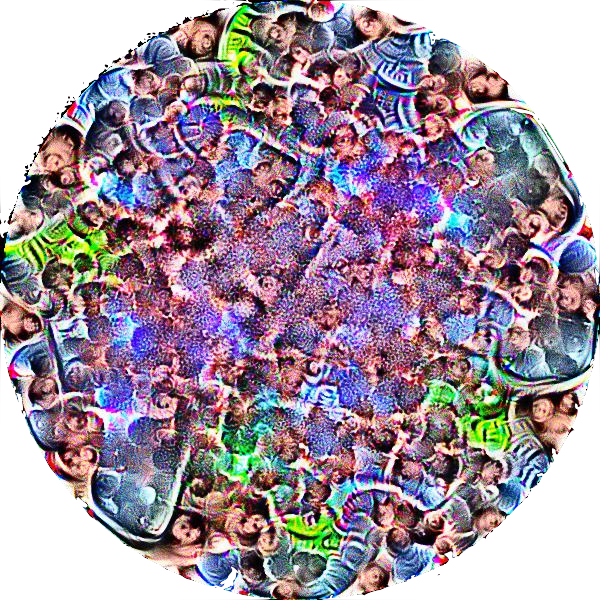}
    \end{center}
    \caption{Circular patches targeting the Person category for FRCNN, SSD, and RetinaNet, respectively.}
    \label{fig:patches}
\end{figure}

By choosing transformations that model different location, scale, and rotation, the optimized patch can remain effective when printed and placed in the real world, where the adversary does not have knowledge of the photographer's position relative to the patch.
Our attacks target 10 object categories from the COCO dataset: apple, backpack, bottle, cat, clock, laptop, microwave, person, potted plant, and suitcase. A patch for each of the 10 categories using both circle and square masks is trained with each of 3 detection models: Faster R-CNN with Resnet-50 \cite{ren2015faster}, SSD with MobileNet \cite{liu2016ssd}, and RetinaNet (SSD with ResNet-50-FPN) \cite{lin2017retina}. In total, we create and print 60 patches for collection. Examples of the trained patches can be seen in Figure \ref{fig:patches}.
Each model is obtained from the TensorFlow Object Detection Model Zoo \cite{huang2017speed}, and comes pre-trained on COCO. We adapt the code provided by Chen et al. \cite{chen2018robust} in order to accommodate single shot models.
All patches are optimized using Expectation over Transformation (Equation \ref{eq:eot}) with the full range of rotations ($0\degree-360\degree$) and with scale ranging from 5 to 50 percent of the image. The Adam optimizer \cite{kingma2015adam} is used with a learning rate of 0.05. The patches are trained for 500 iterations.

Each patch was printed on the high-quality setting of a Canon PRO-100 series printer. The 20 SSD patches are of a different resolution (256x256 pixels) than the Faster R-CNN (FRCNN) and RetinaNet patches (both 600x600 pixels). We scaled all patches to the same size physically, assuming the effects of pixelation are negligible at this size given the collection instruments used. We printed using the Matte Photo Paper setting, printing the entire image on 12 in. x 12 in. non-reflective matte 88lb. 16.5mil (cardstock) paper. The circle patches have a 10 in. diameter, and square patches are 10 in. x 10 in.

\subsection{Dataset Description}

APRICOT comprises 1,011 photos taken by 20 volunteer photographers on cell phones during Summer 2019. Each photo is approximately 12 megapixels. Each photographer was randomly assigned three patches and asked to photograph them in at least five public locations using a variety of distances, angles, indoor and outdoor settings, and photo compositions. The dataset contains between 10 and 42 photos of each patch, with a median of 15.
Each photo in the dataset contains exactly one adversarial patch annotated with a bounding box. The patch generally occupies about 2\% of the image pixels.

Previous work finds that viewing angle affects patch adversariality \cite{chen2018robust}. Because our photos are taken under comparatively uncontrolled settings, we were not able to measure viewing angle at collection time. Instead, we provide three redundant annotations for each photograph that indicate annotators' perception of whether the patch is pictured ``Near Head-On'' (\textless5\degree), ``Angled'' (5\degree - 45\degree), or ``Severely Angled'' (\textgreater45\degree).
Annotators' perceptions are unanimous for approximately 70\% of images.
Our experiments use the mode as the definitive categorization: 
respectively, 39.1\%, 59.5\% and 1.5\% of images are assigned to each category.
One image with no annotator agreement is excluded from analyses that include angle. 
In addition, about 1/4 of the images are labeled as containing a warped patch; i.e., the cardstock the patch is printed on is not flat in the image.

Six patches (138 photos) are assigned to the development set, while the other 54 patches (873 photos) are in the test set. These partitions are photographed by disjoint sets of photographers. The COCO 2017 val and test partitions can be paired with APRICOT for experiments requiring images without adversaries, resulting in approximately 2.5\% adversarial (APRICOT) images per partition. We do not include a training set for fully-supervised models. Annotations are distributed in a format compatible with the COCO API.

\section{Effectiveness of Adversarial Patches} \label{sec:eval}

We present the first systematic analysis of physical adversarial patches in the wild and comparison to digital results. We evaluate the efficacy of our digital adversarial patches by running the COCO-trained detectors on COCO images with the patches inserted (Section \ref{sec:eval_dig}) and on the APRICOT images of physical patches in the real world (Section \ref{sec:eval_phys}). Then we measure the effect of the adversarial patch. 
We define a `fooling event' as a detection that overlaps a ground truth adversarial bounding box, where a `targeted' fooling event is classified as the same object class as the patch's target. 
Performance is reported as the percentage of ground truth adversarial bounding boxes that produced at least one fooling event.

In order to count only predictions caused by the presence of an adversarial patch in the image and not true predictions or non-adversarial model errors, we calculate the intersection over union (IoU) between predicted bounding boxes and ground truth patch bounding boxes and only include predictions with an IoU of at least 0.10. We use a small IoU for our metric because the patches will sometimes generate many small, overlapping predictions in the region of the attack, as can be seen in Figure \ref{fig:patch_example}. These predictions should be preserved because they represent a valid threat to detectors. Still, we find that performance is only slightly degraded by increasing the minimum IoU to 0.50, as most small fooling events are accompanied by larger detections. Some patches overlap larger objects that also create detections, like a patch sitting on a bench. To avoid attributing these to the adversaries, we discard any predicted bounding boxes that are more than twice as large as the ground truth box for the patch. This is necessary because APRICOT is not annotated with COCO object categories.

We report only fooling events with a confidence greater than 0.30. This threshold was chosen because it is the standard threshold used by models in the TensorFlow object detection API model zoo. However, we find that the confidence of fooling event predictions and non-fooling event predictions follow similar distributions. For this evaluation, patches are considered small if they take up less than 10\% of the image, medium if they take up between 10\% and 20\%, and large if they take up greater than 20\%.

\subsection{Digital Performance} \label{sec:eval_dig}

Digital evaluation confirms the attacks' effectiveness before they are inserted into the unconstrained physical world. They also serve as an upper threshold of performance before degradation occurs from real-world effects. Patches have been digitally inserted into random images from the COCO dataset at varying image locations, scales ($5-25 \%$ of the image), and angles of rotation ($0\degree-360\degree$). Each patch was inserted into 15 COCO images in this evaluation.

Table \ref{tbl:res_models} shows the performance of the patches when digitally inserted, broken down by the model used to generate the patch and evaluation model. Results are also broken down by targeted and untargeted fooling events. The patches, for the most part, perform much better when the patch model and evaluation model are the same (white-box attack) than when they are different (black-box attack \cite{papernot2017practical}). For targeted fooling events, white-box performance ranges from $23\%-50\%$, and black-box performance ranges from almost completely ineffective in the worst pairings to nearing white-box performance in the case of FRCNN patches on RetinaNet. We find producing untargeted fooling events to be much easier and see fooling event rates of over $60\%$ even in the black-box case.

We examine the effect of size and shape of the patches on fooling event rate in Table \ref{tbl:blackbox}. The efficacy of adversarial patches increases along with their size in the image in both the targeted and untargeted cases. Square patches perform a little better than circle patches, possibly because patch area was computed using bounding boxes, so circles are consistently over estimated. When measuring untargeted performance, we observe that predicted object categories are clearly correlated with the patch's own geometry; square patches frequently produce kite predictions and circular patches produce doughnut predictions. This highlights the importance of patch shape and is a factor in the difficulty of fooling detectors using patches not tailored to the shape of the targeted category.

\begin{table*}[b]
    \centering
    \resizebox{0.75\linewidth}{!}{
    \begin{tabular}{|c|ll|c|c|c||c|c|c|}
                \cline{4-9}
                \multicolumn{3}{c|}{}& \multicolumn{6}{|c|}{Evaluation Models} \\
                \cline{4-9}
                \multicolumn{3}{c|}{} & \multicolumn{3}{|c||}{Digital}& \multicolumn{3}{|c|}{Physical} \\ \cline{4-9}
                \multicolumn{3}{c|}{} & FRCNN & SSD & RetinaNet & FRCNN & SSD & RetinaNet \\ 
                \hline
          \multirow{8}{1em}{\rotatebox{90}{Patch Models}}& \multirow{2}{4em}{FRCNN (344)} & Targeted & 0.50 & 0.04 & 0.37 & 0.17 & 0.0 & 0.05 \\
          & & Untargeted & 0.89 & 0.64 & 0.84 & 0.43 & 0.15 & 0.34 \\
          \cline{2-9}
         & \multirow{2}{4em}{SSD (331)} & Targeted & 0.08 & 0.23 & 0.07 & 0.14 & 0.06 & 0.04 \\
         & & Untargeted & 0.71 & 0.70 & 0.67 & 0.39 & 0.20 & 0.30 \\
         \cline{2-9}
         & \multirow{2}{5em}{RetinaNet (336)} & Targeted & 0.14 & 0.01 & 0.39  & 0.01 & 0.0 & 0.03 \\
         & & Untargeted & 0.72 & 0.62 & 0.78 & 0.33 & 0.15 & 0.29 \\
         \cline{2-9}
         & \multirow{2}{4em}{All (1,011)} & Targeted & 0.24 & 0.09 & 0.27  & 0.11 & 0.02 & 0.04 \\
         & & Untargeted & 0.77 & 0.65 & 0.76 & 0.38 & 0.17 & 0.31 \\
         \hline
    \end{tabular}
    }
    \captionsetup{width=0.75\linewidth}
    \caption{Performance of digital and physical patches @ 0.3 confidence and 0.10 IoU}
    \label{tbl:res_models}
\end{table*}

\begin{table*}[ht]
    \centering
    \resizebox{\linewidth}{!}{
    \begin{tabular}{|c|l|l|c c |c c |c c|c||c c |c c |c c |c|}
        \cline{4-17}
        \multicolumn{3}{c|}{} & \multicolumn{14}{|c|}{Evaluation Models} \\
        \cline{4-17}
        \multicolumn{3}{c|}{} & \multicolumn{7}{|c||}{Digital} & \multicolumn{7}{|c|}{Physical} \\
        \cline{4-17}
        \multicolumn{3}{c|}{} & \multicolumn{2}{|c|}{FRCNN} & \multicolumn{2}{|c|}{SSD} & \multicolumn{2}{|c|}{RetinaNet} & & \multicolumn{2}{|c|}{FRCNN} & \multicolumn{2}{|c|}{SSD} & \multicolumn{2}{|c|}{RetinaNet} & \\
        \multicolumn{3}{c|}{} & Targeted & Untargeted & Targeted & Untargeted & Targeted & Untargeted & Count & Targeted & Untargeted & Targeted & Untargeted & Targeted & Untargeted & Count\\
        \hline
        \multirow{24}{1em}{\rotatebox{90}{Patch Models}} & \multirow{8}{4em}{FRCNN} & Small & \cellcolor{gray!15}0.07 & \cellcolor{gray!15}0.80 & 0.0 & 0.31 & 0.05 & 0.70 & 100 & \cellcolor{gray!15}0.15 & \cellcolor{gray!15}0.41 & 0.0 & 0.15 & 0.04 & 0.34 & 333\\
        & & Medium & \cellcolor{gray!15}0.64 & \cellcolor{gray!15}0.92 & 0.04 & 0.83 & 0.5 & 0.88 & 120 & \cellcolor{gray!15}1.0 & \cellcolor{gray!15}1.0 & 0.0 & 0.22 & 0.22 & 0.33 & 9\\
        & & Large & \cellcolor{gray!15}0.83 & \cellcolor{gray!15}0.98 & 0.09 & 0.78 & 0.59 & 0.99 & 80 & \cellcolor{gray!15}1.0 & \cellcolor{gray!15}1.0 & 0.0 & 0.0 & 1.0 & 0.0 & 2\\
        \cline{3-17}
        & & Circle & \cellcolor{gray!15}0.47 & \cellcolor{gray!15}0.84 & 0.03 & 0.64 & 0.37 & 0.76 & 150 & \cellcolor{gray!15}0.23 & \cellcolor{gray!15}0.59 & 0.0 & 0.26 & 0.06 & 0.48 & 180\\
        & & Square & \cellcolor{gray!15}0.53 & \cellcolor{gray!15}0.95 & 0.05 & 0.64 & 0.38 & 0.93 & 150 & \cellcolor{gray!15}0.12 & \cellcolor{gray!15}0.26 & 0.0 & 0.04 & 0.04 & 0.19 & 164\\
        \cline{3-17}
        & & Head-On & \cellcolor{gray!15}- & \cellcolor{gray!15}\cellcolor{gray!15}- & - & - & - & - & - & \cellcolor{gray!15}0.21 & \cellcolor{gray!15}0.42 & 0.0 & 0.16 & 0.06 & 0.37 & 126\\
        & & Angled & \cellcolor{gray!15}- & \cellcolor{gray!15}- & - & - & - & - & - &  \cellcolor{gray!15}0.16 & \cellcolor{gray!15}0.42 & 0.0 & 0.14 & 0.04 & 0.31 & 209\\
        & & Severe & \cellcolor{gray!15}- & \cellcolor{gray!15}- & - & - & - & - & - & \cellcolor{gray!15}0.0 & \cellcolor{gray!15}0.78 & 0.0 & 0.44 & 0.0 & 0.56 & 9\\
        \cline{2-17}
        & \multirow{8}{4em}{SSD} & Small & 0.05 & 0.62 & \cellcolor{gray!15}0.03 & \cellcolor{gray!15}0.53 & 0.04 & 0.66 & 110 & 0.11 & 0.34 & \cellcolor{gray!15}0.01 & \cellcolor{gray!15}0.14 & 0.03 & 0.27 & 297\\
        & & Medium & 0.06 & 0.61 & \cellcolor{gray!15}0.22 & \cellcolor{gray!15}0.62 & 0.09 & 0.62 & 110 & 0.32 & 0.64 & \cellcolor{gray!15}0.41 & \cellcolor{gray!15}0.63 & 0.0 & 0.50 & 22\\
        & & Large & 0.14 & 0.97 & \cellcolor{gray!15}0.51 & \cellcolor{gray!15}0.92 & 0.10 & 0.78 & 80 & 0.33 & 0.83 & \cellcolor{gray!15}0.67 & \cellcolor{gray!15}1.0 & 0.33 & 0.83 & 12\\
        \cline{3-17}
        & & Circle & 0.08 & 0.59 & \cellcolor{gray!15}0.20 & \cellcolor{gray!15}0.70 & 0.05 & 0.58 & 150 & 0.09 & 0.35 & \cellcolor{gray!15}0.01 & \cellcolor{gray!15}0.14 & 0.03 & 0.31 & 156\\
        & & Square & 0.07 & 0.82 & \cellcolor{gray!15}0.25 & \cellcolor{gray!15}0.70 & 0.10 & 0.77 & 150 & 0.18 & 0.42 & \cellcolor{gray!15}0.10 & \cellcolor{gray!15}0.25 & 0.05 & 0.30 & 175\\
        \cline{3-17}
        & & Head-On & - & - & \cellcolor{gray!15}- & \cellcolor{gray!15}- & - & - & - & 0.17 & 0.43 & \cellcolor{gray!15}0.06 & \cellcolor{gray!15}0.20 & 0.04 & 0.26 & 141 \\
        & & Angled & - & - & \cellcolor{gray!15}- & \cellcolor{gray!15}- & - & - & - & 0.11 & 0.35 & \cellcolor{gray!15}0.06 & \cellcolor{gray!15}0.19 & 0.03 & 0.30 & 188\\
        & & Severe & - & - & \cellcolor{gray!15}- & \cellcolor{gray!15}- & - & - & - & 0.0 & 0.0 & \cellcolor{gray!15}0.0 & \cellcolor{gray!15}1.0 & 0.0 & 0.0 & 1\\
        \cline{2-17}
        & \multirow{8}{5em}{RetinaNet} & Small & 0.02 & 0.77 & 0.0 & 0.35 & \cellcolor{gray!15}0.05 & \cellcolor{gray!15}0.61 & 100 & 0.02 & 0.33 & 0.0 & 0.13 & \cellcolor{gray!15}0.01 & \cellcolor{gray!15}0.27 & 317\\
        & & Medium & 0.18 & 0.65 & 0.02 & 0.77 & \cellcolor{gray!15}0.48 & \cellcolor{gray!15}0.77 & 120 & 0.0 & 0.31 & 0.0 & 0.38 & \cellcolor{gray!15}0.38 & \cellcolor{gray!15}0.51 & 16\\
        & & Large & 0.24 & 0.78 & 0.03 & 0.77 & \cellcolor{gray!15}0.60 & \cellcolor{gray!15}0.94 & 80 & 0.0 & 1.0 & 0.0 & 1.0 & \cellcolor{gray!15}0.0 & \cellcolor{gray!15}1.0 & 3\\
        \cline{3-17}
        & & Circle & 0.17 & 0.70 & 0.01 & 0.64 & \cellcolor{gray!15}0.35 & \cellcolor{gray!15}0.70 & 150 & 0.03 & 0.40 & 0.0 & 0.21 & \cellcolor{gray!15}0.06 & \cellcolor{gray!15}0.37 & 170\\
        & & Square & 0.11 & 0.74 & 0.02 & 0.61 & \cellcolor{gray!15}0.39 & \cellcolor{gray!15}0.82 & 150 & 0.0 & 0.27 & 0.0 & 0.09 & \cellcolor{gray!15}0.0 & \cellcolor{gray!15}0.21 & 166\\
        \cline{3-17}
        & & Head-On & - & - & - & - & \cellcolor{gray!15}- & \cellcolor{gray!15}- & - & 0.03 & 0.37 & 0.0 & 0.16 & \cellcolor{gray!15}0.02 & \cellcolor{gray!15}0.25 & 127\\
        & & Angled & - & - & - & - & \cellcolor{gray!15}- & \cellcolor{gray!15}- & - & 0.01 & 0.32 & 0.0 & 0.15 & \cellcolor{gray!15}0.03 & \cellcolor{gray!15}0.30 & 204\\
        & & Severe & - & - & - & - & \cellcolor{gray!15}- & \cellcolor{gray!15}- & - & 0.0 & 0.2 & 0.0 & 0.0 & \cellcolor{gray!15}0.0 & \cellcolor{gray!15}0.4 & 5\\
        \hline
    \end{tabular}
    }
    \caption{Black-box and white-box performance of digital and physical patches broken down by size, shape, and angle. Angle is only taken into account for physical patches. Grayed boxes correspond to white-box case. Count columns indicate the number of images used for evaluation.}
    \label{tbl:blackbox}
\end{table*}

\subsection{Physical Performance} \label{sec:eval_phys}

Table \ref{tbl:res_models} compares physical and digital patch performance in the white-box and black-box cases. In the physical domain, the division of patch size and angle is not evenly distributed, as this is difficult to ensure when people are taking photographs. The efficacy of patches decreases when photographed in the real world. We observe many of the same trends as in the digital results: untargeted fooling events are much easier to produce and certain models are easier to fool in both the white-box and black-box cases. The correlation between digital and physical results indicates that an adversary can develop an attack digitally and have some confidence as to its efficacy in the real world. However, these results also reveal that fooling detectors in the wild is challenging, especially given black-box knowledge.

As in the digital evaluation, the physical performance of the patches is analysed with respect to several patch characteristics: shape, size, and, additionally, angle to the camera. Comparison to the digital results can be found in Table \ref{tbl:blackbox}. Again, larger patches have a higher fooling rate for all models in targeted and untargeted cases. Circular patches appear to be more effective on FRCNN and RetinaNet, but square patches are more effective on SSD. As in the digital evaluation, patch geometry has an effect on untargeted category predictions. Circular patches favor kite and umbrella, while also producing frisbee, sports ball, and doughnut. In contrast, square patches favor categories such as TV, handbag, book, and kite. We hypothesize kite is a common prediction due to sharing bright colors with the patches. As angle becomes more extreme, targeted performance decreases. Note that shear transformations were not used during patch training. Interestingly, in the untargeted case, head-on and angled patches perform roughly the same, but severely angled patches perform best by a large margin. Untargeted predictions for severely angled patches favor categories such as kite and umbrella, and also include surfboard and laptop.

Some of the patches included in APRICOT contain regions that closely resemble the target object class. For example, apple-like patterns can be seen in the patch in Figure \ref{fig:patch_example}a. This phenomenon was also observed by \cite{chen2018robust}. This result is not entirely surprising, as the optimization framework of \cite{chen2018robust} resembles \cite{simonyan2013deep} which instead used gradient ascent as a tool for network explanation and object class visualization. Even though some patches resemble natural objects, they may still be considered adversarial if they possess an unnatural hyper-objectness characteristic. The work of \cite{brown2017adversarial} was able to produce adversarial patches that influence a classifier's output more strongly than a genuine object instance. This shows that adversarial patterns can manipulate a network's internal representations in an unnatural way to draw more attention than a natural object. This phenomenon could explain why the patches in APRICOT trigger so many untargeted detections even when severely angled.

\section{Flagging Adversarial Patches} \label{sec:detection}

Detection systems that can accurately flag adversaries could neutralize their risk by identifying and filtering malicious results.
There are multiple ways to formulate the task of flagging adversaries. One could train a detector to directly output predictions for patch locations. One could also apply defensive techniques to a detection network's outputs to identify if one of its predictions was caused by an adversary. Or one could ignore the detector and directly assess image regions to predict whether they contain an adversarial patch. We present baseline experiments to address each of these strategies.

We consider both supervised and unsupervised defenses. In this context, ``supervised defense'' means that we assume the defender has some knowledge of the attacker's algorithm and thus can sample their own patches from the same or similar distribution and use them to train a defensive mechanism. In the digital defense domain, this is analogous to \cite{metzen2017detecting}, however it has not yet been studied if robust physical adversarial patches can be easily detected this way. Even if such a defense works, it may not be sufficient if it fails to generalize to new attack strategies, as was observed in \cite{metzen2017detecting}.
In contrast, ``unsupervised defenses'' have no prior knowledge of the attack strategy, and instead must learn to identify adversarial anomalies after training on only normal data. This is clearly more challenging, and we believe it presents a more realistic defense scenario as new attacks are always being developed. These approaches are completely attack-agnostic but may simply expose anomalies in the data without regard for adversariality.
Results are reported on the APRICOT test partition.

\subsection{Detecting Adversarial Patches using Synthetic Supervision} \label{sec:fp}

\begin{figure}
\centering
\begin{minipage}{.5\linewidth}
  \centering
  \captionsetup{width=0.9\linewidth}
  \includegraphics[width=0.9\linewidth]{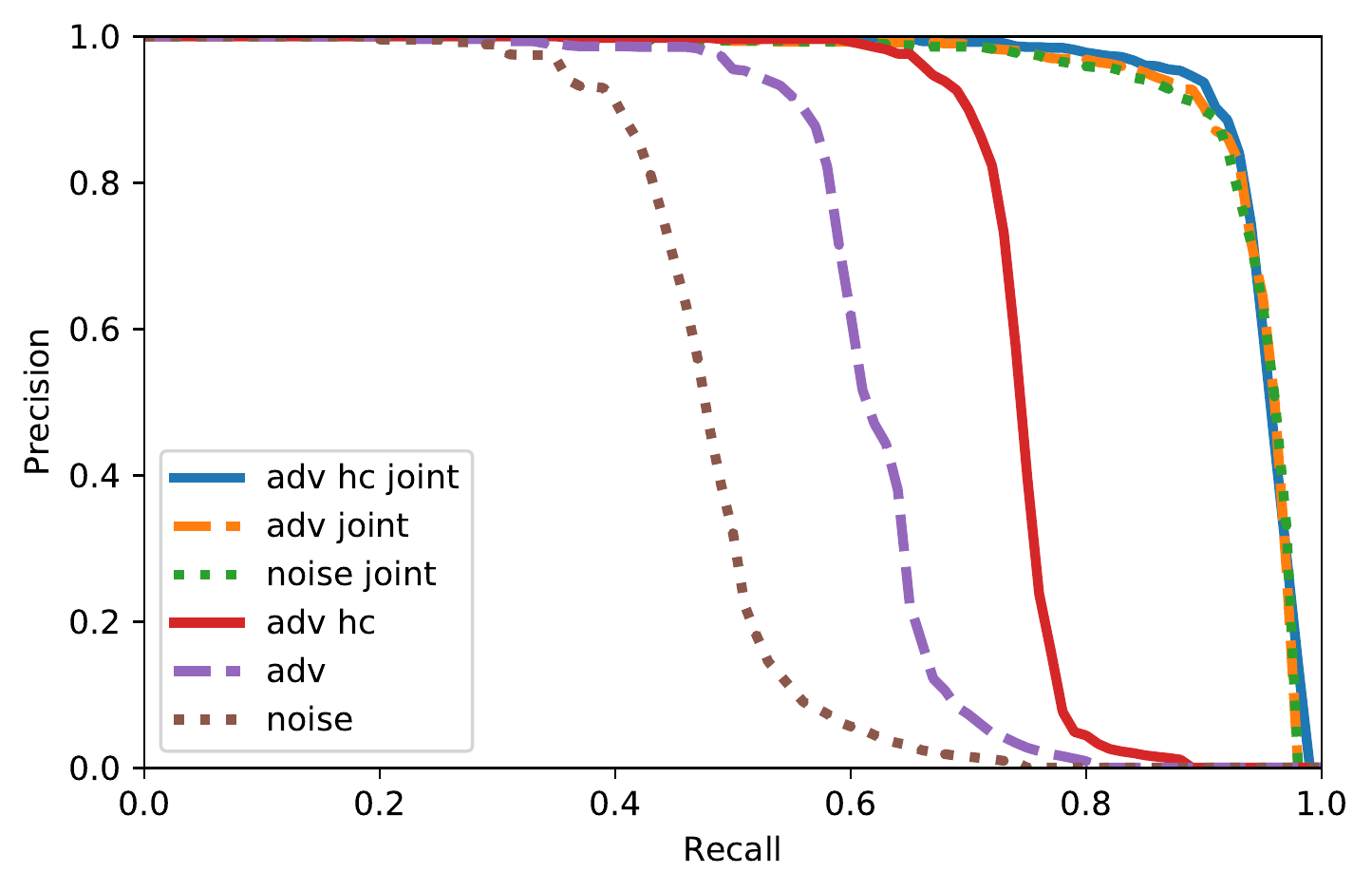}
  \captionof{figure}{Precision-Recall curves on APRICOT for models trained on synthetic flying patch images}
  \label{fig:fp_pr}
\end{minipage}\begin{minipage}{.5\linewidth}
  \centering
  \captionsetup{width=0.9\linewidth}
  \includegraphics[width=0.9\linewidth]{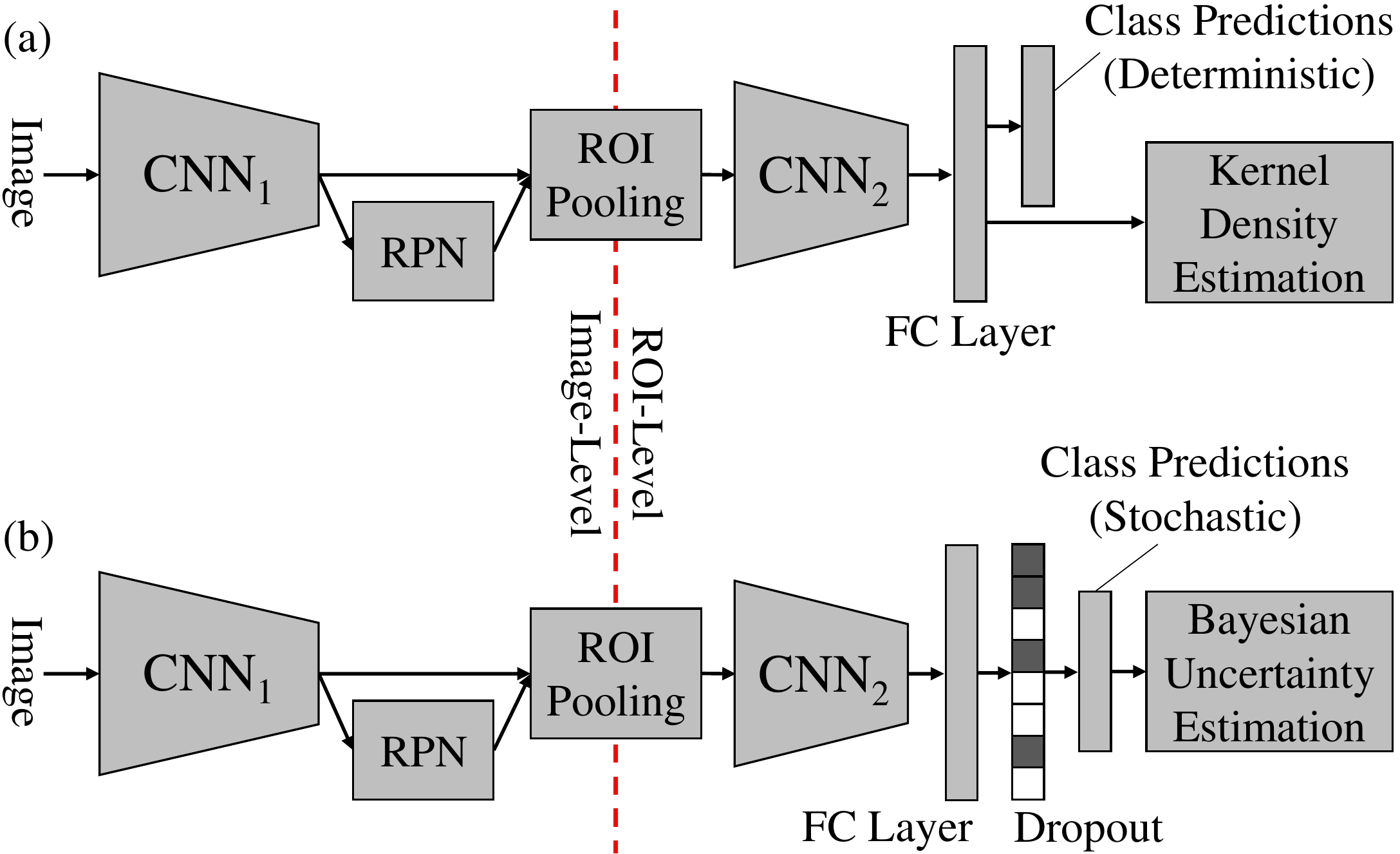}
  \captionof{figure}{Adversarial detection strategies of \cite{feinman2017detecting} modified for Faster R-CNN. (a) Shows the layer used for KDE. (b) A dropout layer is added for Bayesian uncertainty estimation.}
  \label{fig:uncertainty}
\end{minipage}
\end{figure}

We train a Faster R-CNN \cite{ren2015faster} object detection model to detect adversarial patches using a synthetic dataset of patches digitally overlaid on the COCO 2017 training set. APRICOT does not contain a training partition, and the cost of physical data collection makes it impractical to collect an appropriately large dataset for fully-supervised adversary detector training. Instead we create several synthetic Flying Patch Datasets (named in homage to \cite{dosovitskiy2015flownet}) by overlaying patches of several shapes in random positions, sizes, and angles on natural images, similar to the method used in our digital performance analysis. Our adversarial Flying Patch Datasets also use the method of \cite{chen2018robust} to create patches. Under this framework, patches have a confidence parameter which limits their intensity through L2 regularization. We train models on two datasets, one with a range of confidence settings similar to \cite{chen2018robust}, and one with only \textit{high confidence} patches with the L2 regularization disabled. A dataset of Gaussian noise patches is used as a control to examine whether the models learn features of the attack patterns or more superficial features such as shape and color. For each flying patch collection, two models are trained. The  \textit{joint} variant detects both patches and COCO objects, while \textit{patch only} models detect only adversarial patches.

Our patch detector models are fine-tuned from a publicly available COCO detection model, which is the same model used in \cite{chen2018robust}. This model is different from any used to create patches in APRICOT, so all attacks in APRICOT would be considered black-box attacks.
Table \ref{tbl:fp_res} reports patch detection performance as Average Precision (AP) and Average Recall (AR). We report results under the default settings of the COCO API \cite{lin2014microsoft}: maximum detections 100 and IoU threshold ranging from 0.5 to 0.95. In practice, if an automated system is able to correctly detect a sub-region of an adversarial patch, this detection should be sufficient to alert the system to the presence of the patch. For this reason, we also report AP and AR under a more lenient IoU threshold of 0.5. We also provide a breakdown of recall by target model for the APRICOT patches. Figure \ref{fig:fp_pr} visualizes precision and recall with a 0.5 IoU threshold and max 100 detections.

We find that synthetic data is sufficient to train a high-performing adversarial patch detector for exemplars in the physical world. For all training configurations, performance is significantly improved by jointly training to detect both patches and normal objects. This suggests that modeling normal objects reduces false positives for adversaries. The best performing model is the one trained on high confidence adversarial patches; however, under a lenient IoU threshold, the performance difference between the jointly trained models becomes much less significant. This suggests that patch shape, color, and texture intensity provide enough cues to detect adversarial patches. Error analysis of high-scoring false positives in APRICOT and COCO reveals mostly colorful, high-texture objects.

While these supervised models achieve high average precision, there is a risk they will fail to generalize to new attacks like in \cite{metzen2017detecting}. This motivates the need for attack-agnostic unsupervised defenses like \cite{samangouei2018defense,song2018pixeldefend}. In addition, clever attackers could likely evade detection with white-box access to the patch detector, or by using regularizations that create patches closer to the color and texture distributions of natural scenes.

\subsection{Determining if a Detection is Adversarial using Uncertainty and Density}

We adapt the approaches of \cite{feinman2017detecting} to determine whether we can discriminate detections produced by adversarial patches from normal predictions made by an object detection model. This method is a self-defense for the object detection model and does not address adversaries that fail to produce false positives. These approaches examine all predictions made by the same model as Section \ref{sec:fp} with \textgreater0.3 confidence, resulting in 273 untargeted fooling events out of 4428 detections in APRICOT's test partition.
The results are therefore described using retrieval statistics over the 4428 model detections, and these methods' results are tied to the particular detection model's accuracy and internal representations.

\begin{table*}[tp]
    \centering
    \resizebox{0.75\linewidth}{!}{
    \begin{tabular}{|l|l|l|l||l|l|l|l|l|}
        \cline{3-9}
        \multicolumn{2}{c|}{} & \multicolumn{2}{c||}{IoU$_{0.5:0.95}$} & \multicolumn{5}{c|}{IoU$_{0.5}$} \\
        \hline
        Objective & Training Patches & AP & AR & AP & AR & AR$_{frcnn}$ & AR$_{ret}$ & AR$_{ssd}$  \\ 
        \hline
        \multirow{3}{5em}{Joint} & Adv. High Conf. & \textbf{0.686} & \textbf{0.757} & \textbf{0.943} & \textbf{0.981} & \textbf{1.000} & 0.941 & \textbf{1.000} \\
        & Adversarial & 0.670 & 0.749 & 0.937 & 0.976 & 0.983 & 0.944 & \textbf{1.000} \\
        & Gaussian & 0.635 & 0.714 & 0.936 & 0.979 & 0.987 & \textbf{0.958} & 0.993 \\
        \hline
        \multirow{3}{5em}{Patch Only} & Adv. High Conf. & 0.540 & 0.655 & 0.739 & 0.885 & 0.893 & 0.785 & 0.979 \\
        & Adversarial & 0.447 & 0.588 & 0.615 & 0.805 & 0.759 & 0.691 & 0.969 \\
        & Gaussian & 0.329 & 0.500 & 0.478 & 0.742 & 0.696 & 0.562 & 0.872 \\
        \hline
    \end{tabular}
    }
    \captionsetup{width=0.75\linewidth}
    \caption{Performance on APRICOT of detector models trained on synthetic flying patch images}
    \label{tbl:fp_res}
\end{table*}

Figure \ref{fig:uncertainty} illustrates how embeddings and uncertainty are extracted for the Faster R-CNN model. The density approach learns a Kernel Density Estimation (KDE) model for each output class. In this case, we learn a KDE over the final latent representation passed to the detector's object classifier, for all detections in  COCO val 2017 with at least 0.3 confidence. The embedding for each detection in APRICOT is scored with respect to the detected class.
The Bayesian uncertainty approach of \cite{feinman2017detecting} is considered to be more robust to adversarial circumvention \cite{carlini2017adversarial}. We adapt this approach by applying 0.5 dropout to the same Faster R-CNN layer at inference time and measuring variance in resulting object category predictions. It would not be possible to add dropout before the region of interest (ROI) pooling step, as each random trial would alter the region proposals themselves. Performing dropout at only the second to last layer is also computationally efficient, as added random trials incur minimal cost. Scores from these two methods are then used as features in a logistic regression learned from the APRICOT dev set, requiring a small amount of supervision. Note that the individual metrics require no supervision.

Figure \ref{fig:kde_all} shows Receiver Operating Characteristics for these three approaches under a low-information scenario where a single threshold is set for all object classes. All three approaches are effective for this task. There is insufficient data to set individual thresholds per detected class, as would be most appropriate for the KDE models; only four object classes have at least 10 fooling events under this model. Many fooling events do not produce the targeted class prediction; this suggests that uncertainty is particularly effective because these detections lie near decision boundaries. The additional supervision required to produce the combined model is only marginally beneficial over uncertainty alone. These results are confounded by non-adversarial false positives or unlikely exemplars of COCO categories. More investigation is needed to determine if there is a difference in how true and false positives are scored in these analyses.

\subsection{Localizing Adversarial Regions with Density and Reconstruction}

Experiments using autoencoders localize anomalous regions without regard for any attack strategy or target model. These tests are conducted on (32,32) windows of images. We model natural images with the ACAI autoencoder described in \cite{berthelot*2018understanding}, trained on windows of the images from COCO's 2017 train partition. Before constructing the windows, all images are resized to (600,600).
Unlike the detection models in \ref{sec:fp}, these localization results label small, uniformly gridded windows of the image that are not necessarily aligned to patch bounding boxes. Therefore, they are presented using receiver operator characteristics over small windows. In this experiment, a (32,32) pixel window is considered adversarial if at least half of the pixels are contained in an APRICOT bounding box.

\begin{figure}
\centering
\begin{minipage}{.5\linewidth}
  \centering
  \captionsetup{width=0.9\linewidth}
  \includegraphics[width=0.9\linewidth]{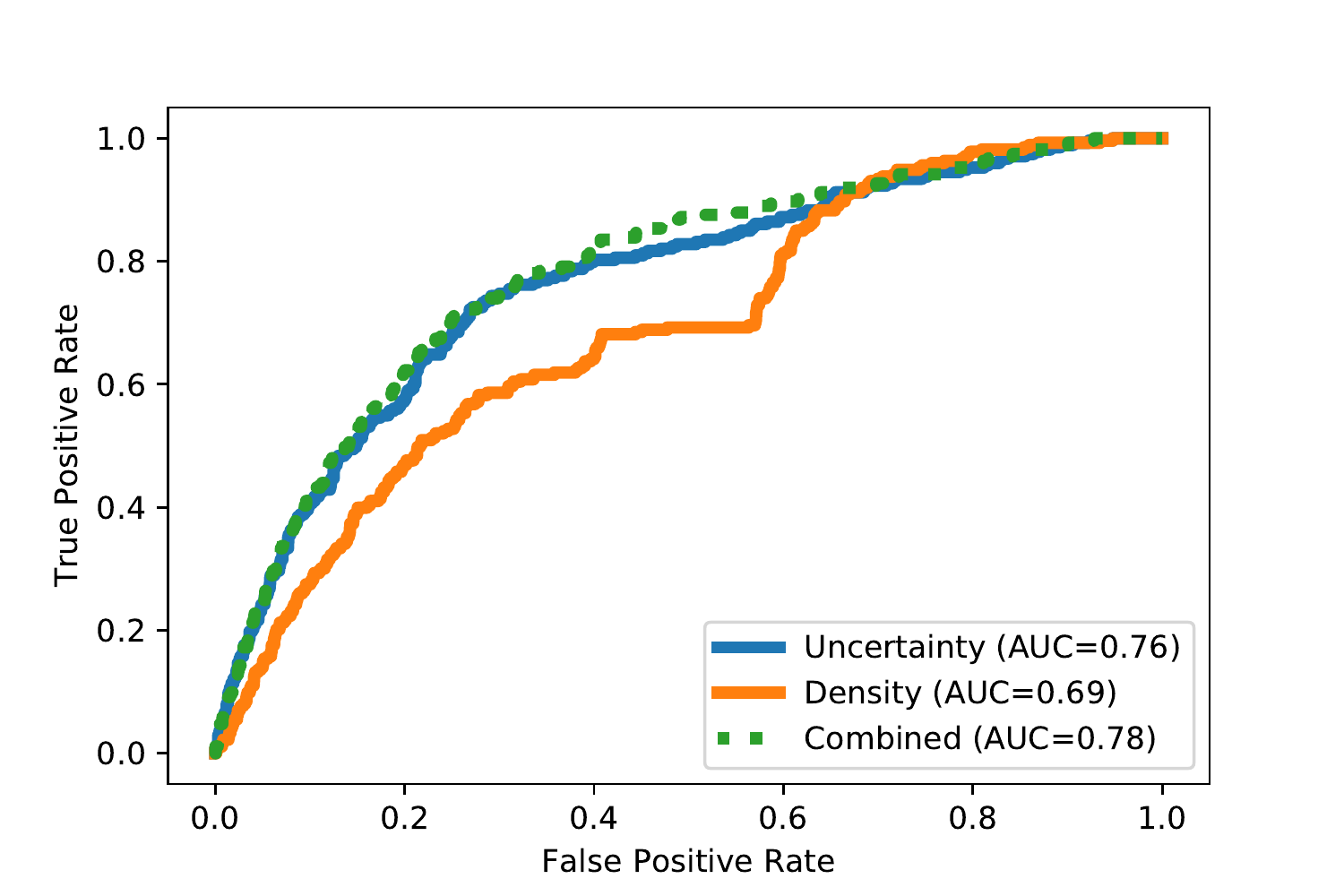}
  \captionof{figure}{ROC of KDE and Bayesian uncertainty for finding adversarial detections}
  \label{fig:kde_all}
\end{minipage}\begin{minipage}{.5\linewidth}
  \centering
  \captionsetup{width=0.9\linewidth}
  \includegraphics[width=0.9\linewidth]{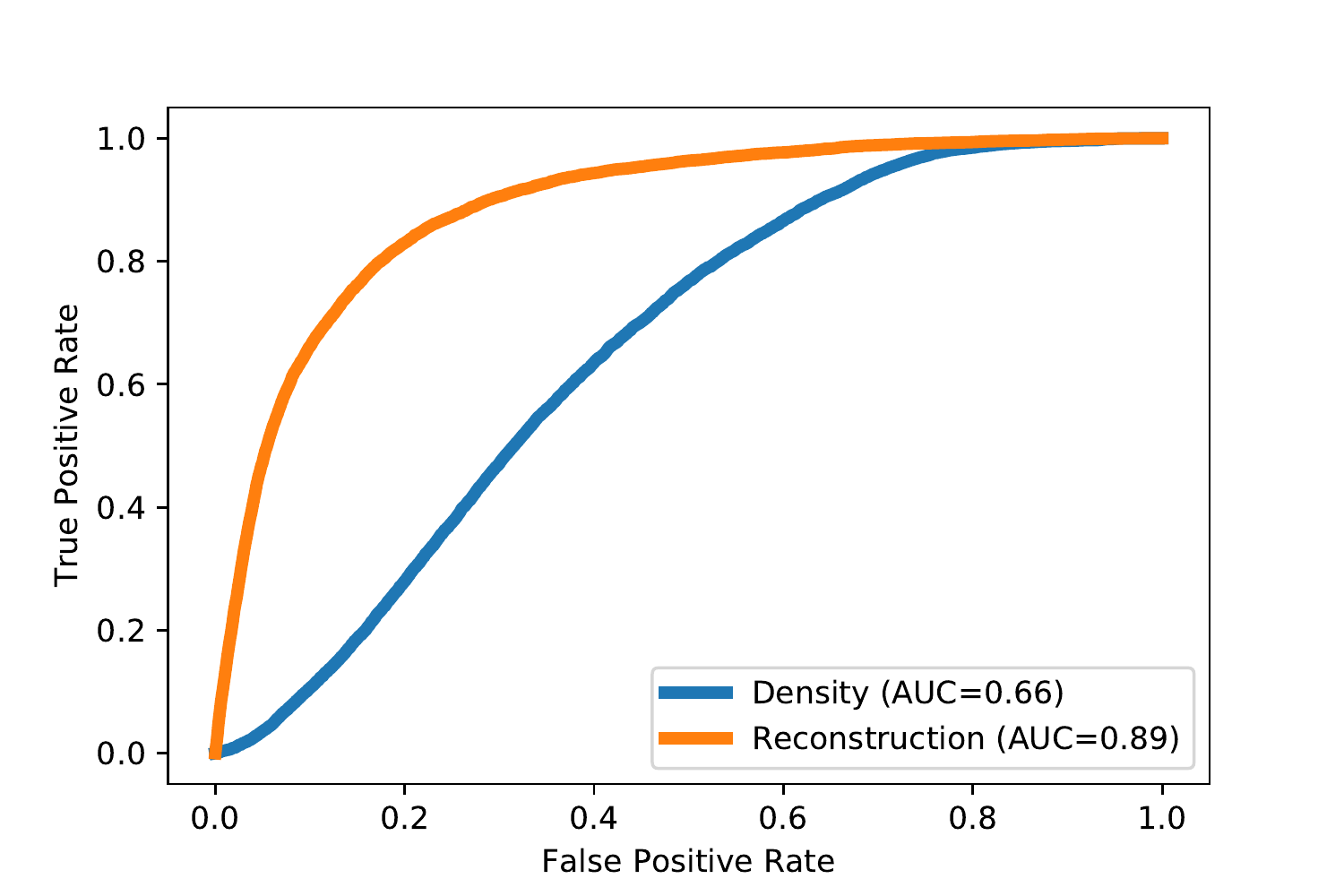}
  \captionof{figure}{Performance of autoencoder methods for localizing adversary-containing 32x32 image regions}
  \label{fig:window_detect}
\end{minipage}
\end{figure}

The density approach is further adapted to learn a one class model over embeddings of the non-adversarial data in the COCO 2017 validation partition. A Gaussian Mixture Model is employed on the intuition that since the embeddings are not separated by class, they will be multimodal. Five components were chosen empirically. Each image window is represented by mean-pooling the spatially invariant dimensions of the autoencoder bottleneck.

A previous work \cite{schlegl2017unsupervised} used GAN inversion and reconstruction error as a tool for anomaly detection in medical imagery. In the context of APRICOT, adversarial patches may be detectable as anomalies. Another work \cite{samangouei2018defense} used GANs trained on normal data as a tool to remove adversarial perturbations from digital attacks. However, both of these GAN-based reconstruction methods rely on expensive computations to recover the latent space vector and reconstruct the image. We reconstruct each window instead using the autoencoder and measure the L2 distance of the reconstruction from the original.

As shown in Figure \ref{fig:window_detect}, the density-based approach has weak predictive power, while reconstruction error seems to effectively discriminate adversarial windows. High-frequency natural textures like foliage generate a high reconstruction error, producing false positives. These differences in performance suggest that the discriminative power of the autoencoder comes not from the information encoded in the latent space but rather from the information lost during encoding.

\section{Conclusion}

Research into the effectiveness and detectability of adversarial objects in the wild is limited by the absence of labeled datasets of such objects in the real world. We provide the first publicly available dataset of physical adversarial patches in a variety of real-world settings. Annotations of the patches' locations, attacked model, warping, shape, and angle to the camera are included to allow investigation into the effects of these attributes on the patches' adversariality.

Our analysis reveals that photographed physical patches in the wild can fool both white-box and, in some cases, black-box object detection models. Benchmark flagging experiments establish that adversarial patches can be flagged both with and without prior knowledge of the attacks. While our proposed baseline defenses seem to perform well on APRICOT, these efforts should not be considered the answer; in industry, even a small percentage of error can mean the difference between success and catastrophic failure. Moving forward, we want to encourage further experimentation to increase detection accuracy as adversarial attacks continue to become more and more complex.

\section*{Acknowledgments}
We would like to thank Mikel Rodriguez, David Jacobs, Rama Chellappa, and Abhinav Shrivastava for helpful discussions and feedback on this work. We would also like to thank our MITRE colleagues who participated in collecting and annotating the APRICOT dataset and creating the adversarial patches.

%
%
\bibliographystyle{splncs04}
\bibliography{egbib}


\noindent
\begin{figure*}[ht!]
  \centering
    \includegraphics[width=\textwidth]{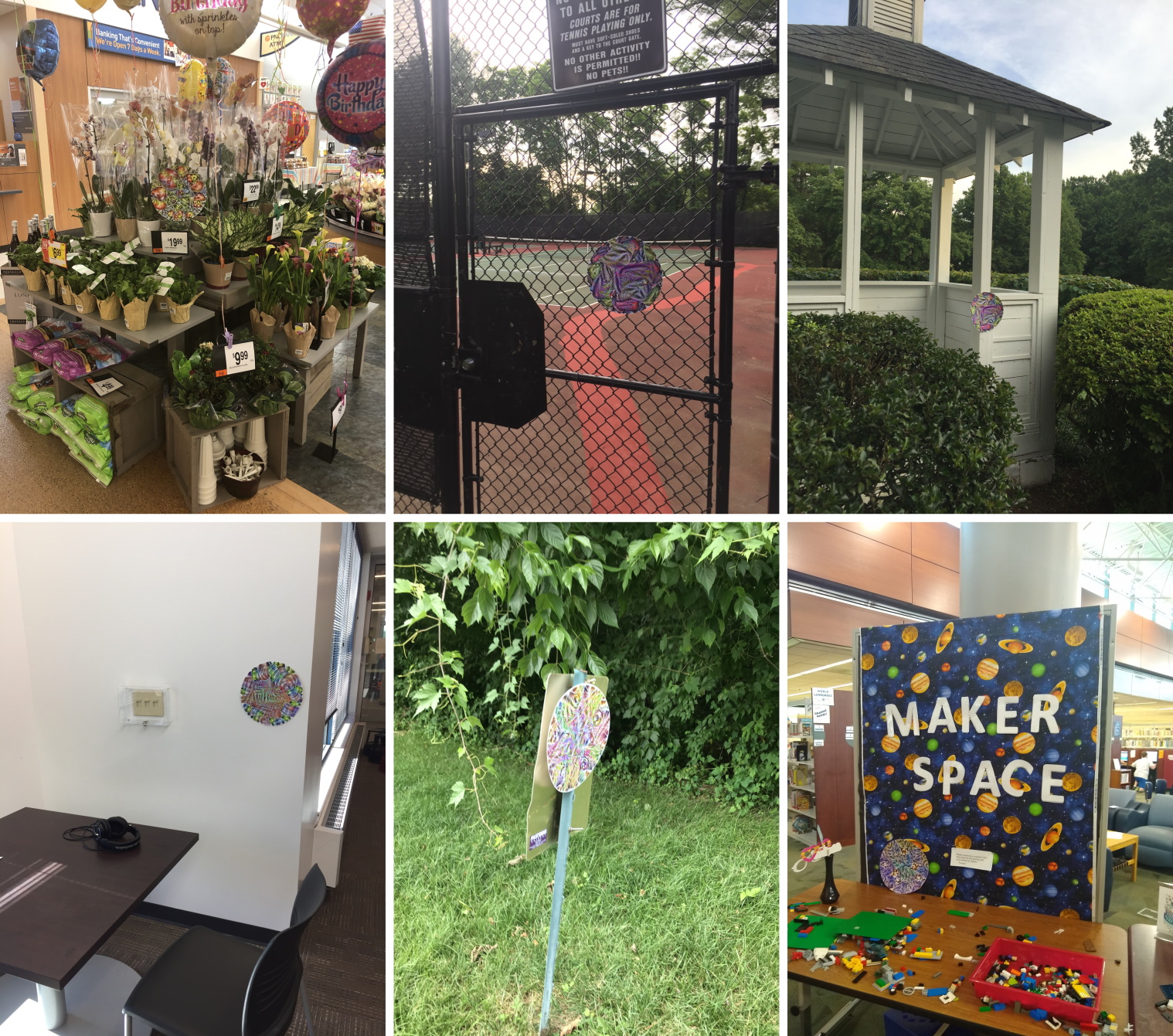}
  \caption{Sample images from the APRICOT dataset. Images have been downsampled to reduce file size.}
  \label{fig:extra1}
\end{figure*}

\begin{figure*}[t]
  \centering
    \includegraphics[width=\textwidth]{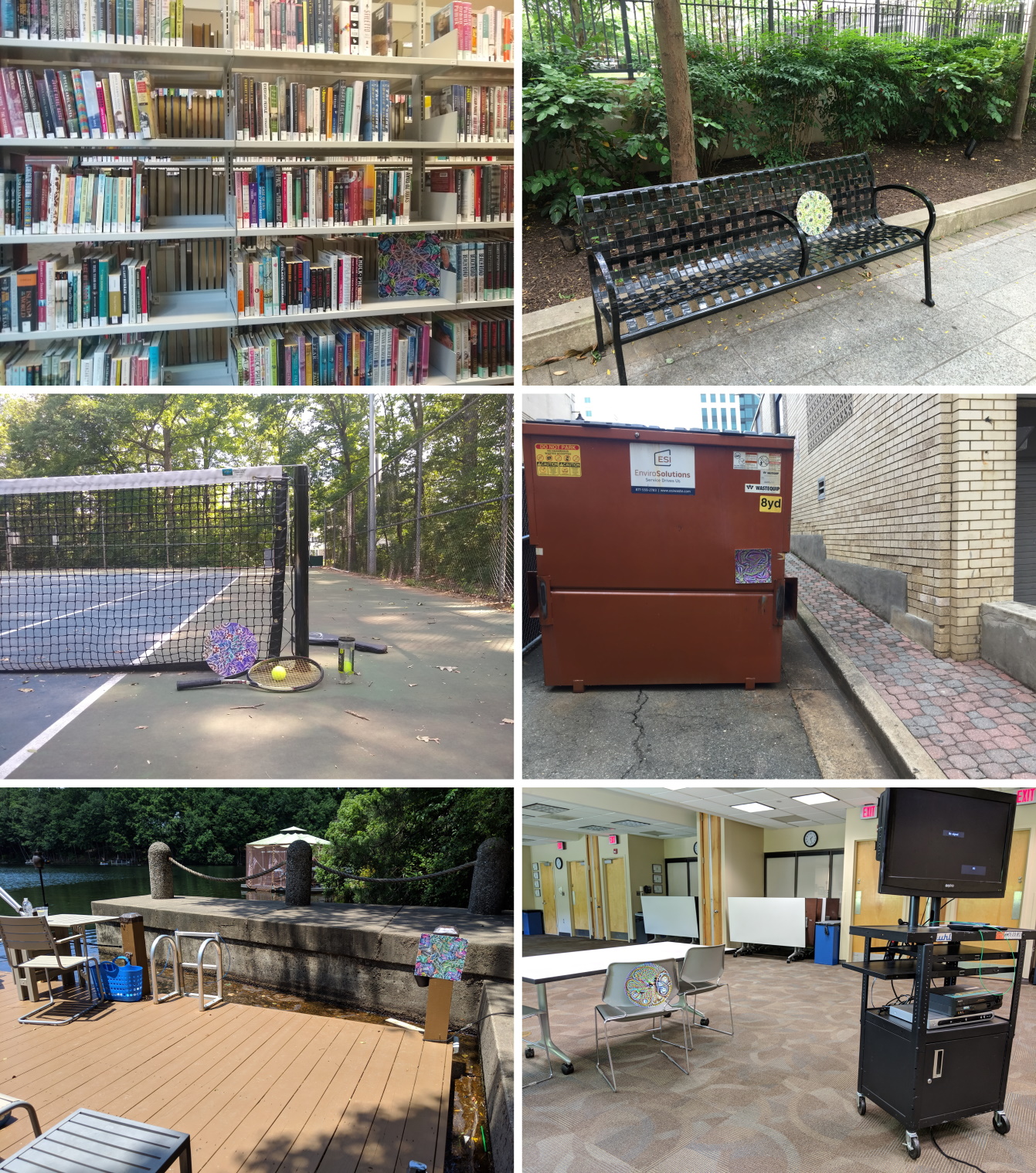}
  \caption{Sample images from the APRICOT dataset}
  \label{fig:extra1}
\end{figure*}

\begin{figure*}[t]
  \centering
    \includegraphics[width=\textwidth]{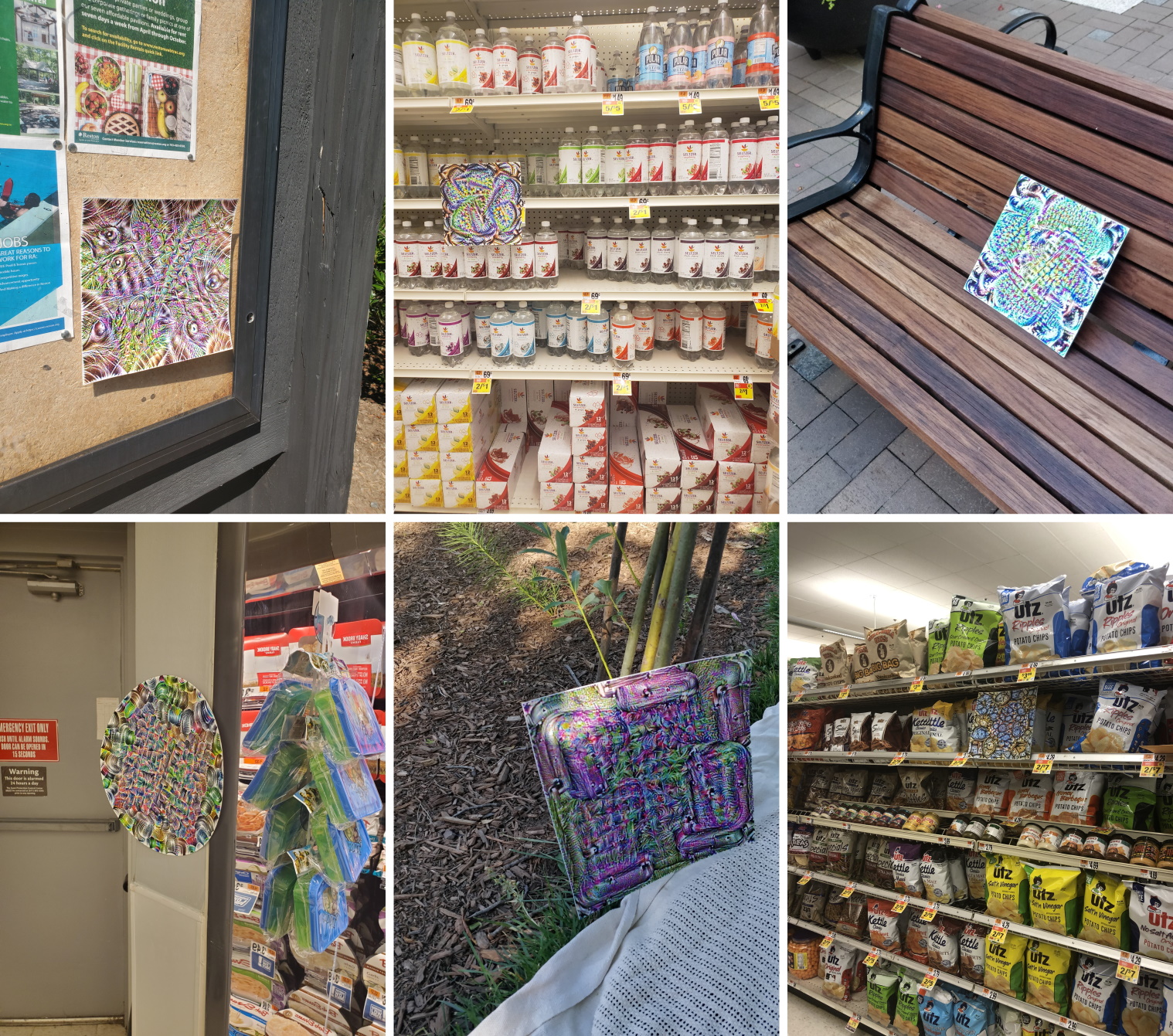}
  \caption{Sample images from the APRICOT dataset}
  \label{fig:extra1}
\end{figure*}

\begin{figure*}[t]
  \centering
    \includegraphics[width=\textwidth]{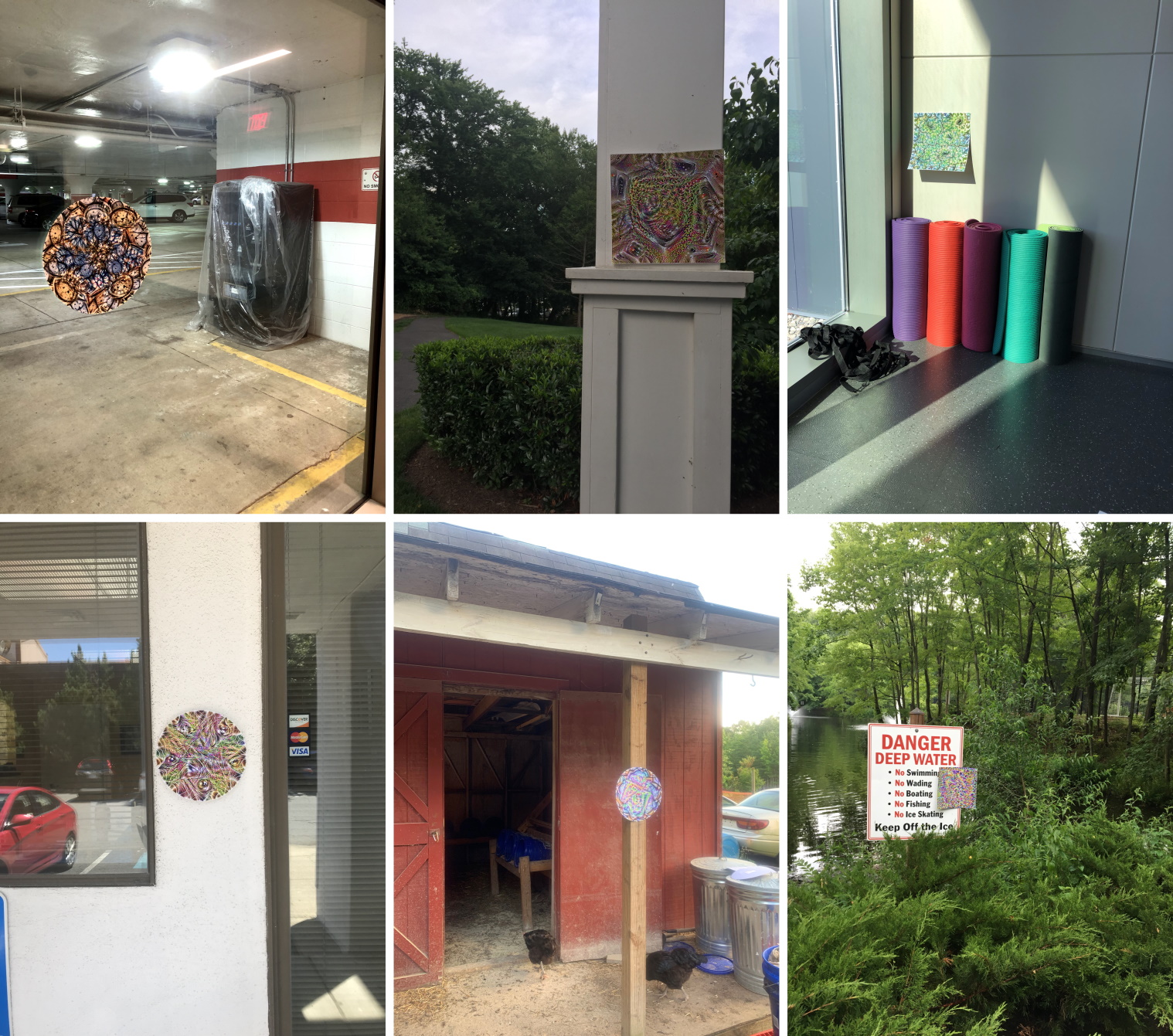}
  \caption{Sample images from the APRICOT dataset}
  \label{fig:extra1}
\end{figure*}


\begin{figure}[t]
  \centering
    \includegraphics[width=0.5\textwidth]{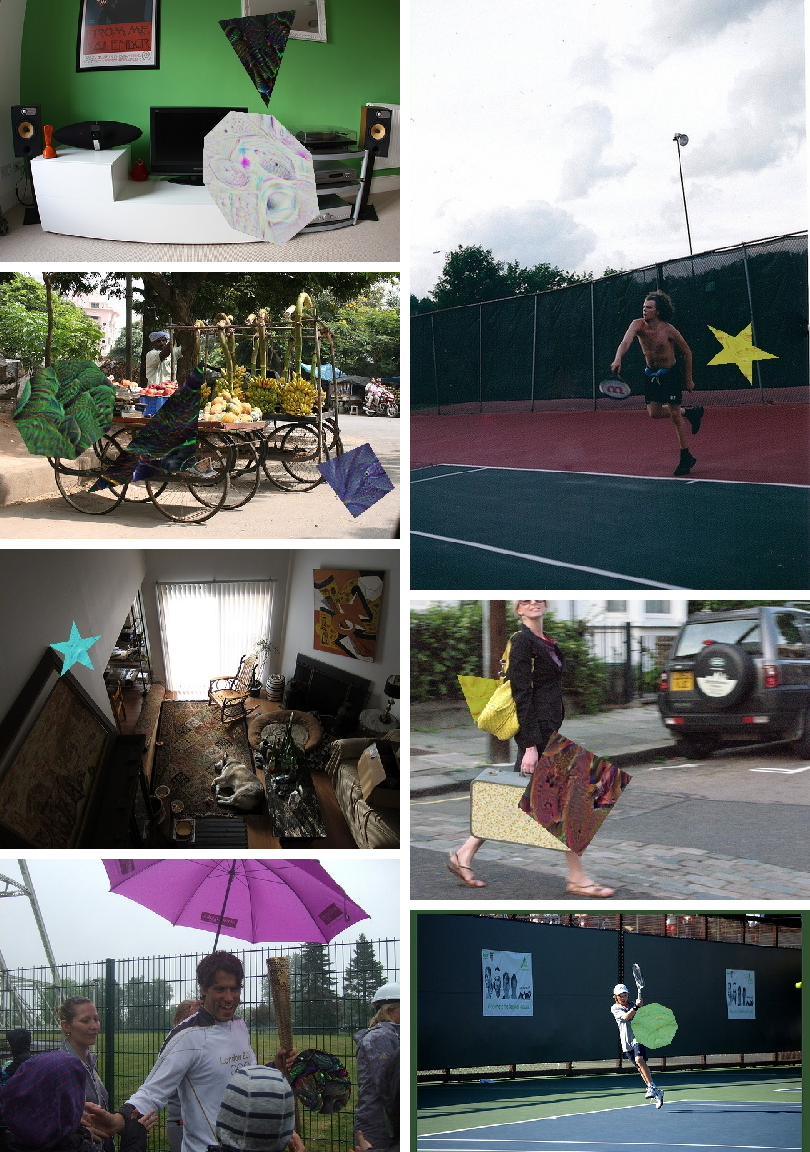}
  \caption{Sample of ``flying patch'' images with digital patches superimposed over COCO images, which were used to train adversarial patch detectors}
  \label{fig:fpsamp}
\end{figure}

\begin{figure}
  \centering
    \includegraphics[width=0.5\textwidth]{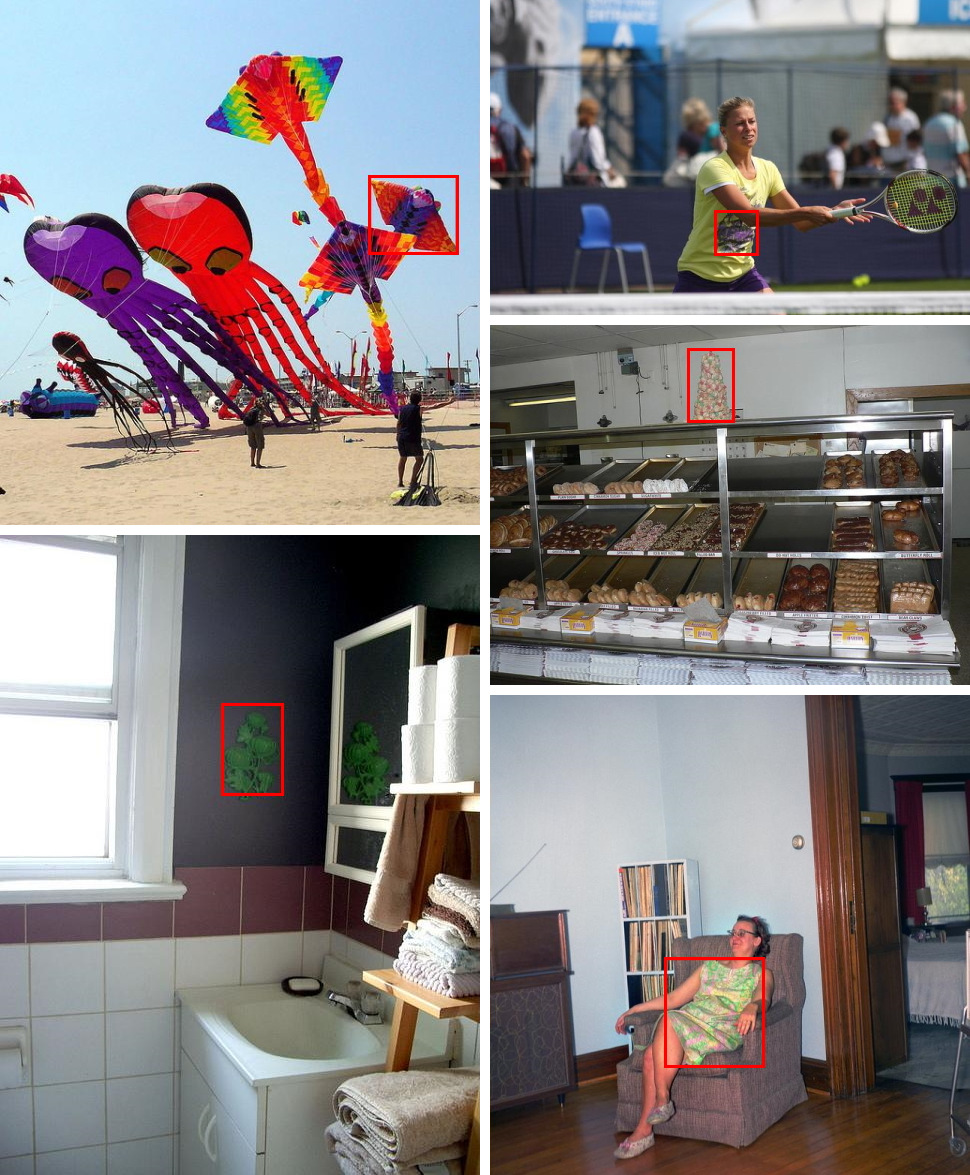}
  \caption{Sample of high-scoring false positive patch detections in the COCO 2017 test set as produced by the best-performing model trained on flying patch images}
  \label{fig:fp_false_pos}
\end{figure}

\begin{figure}[t]
  \centering
    \includegraphics[width=1.0\textwidth]{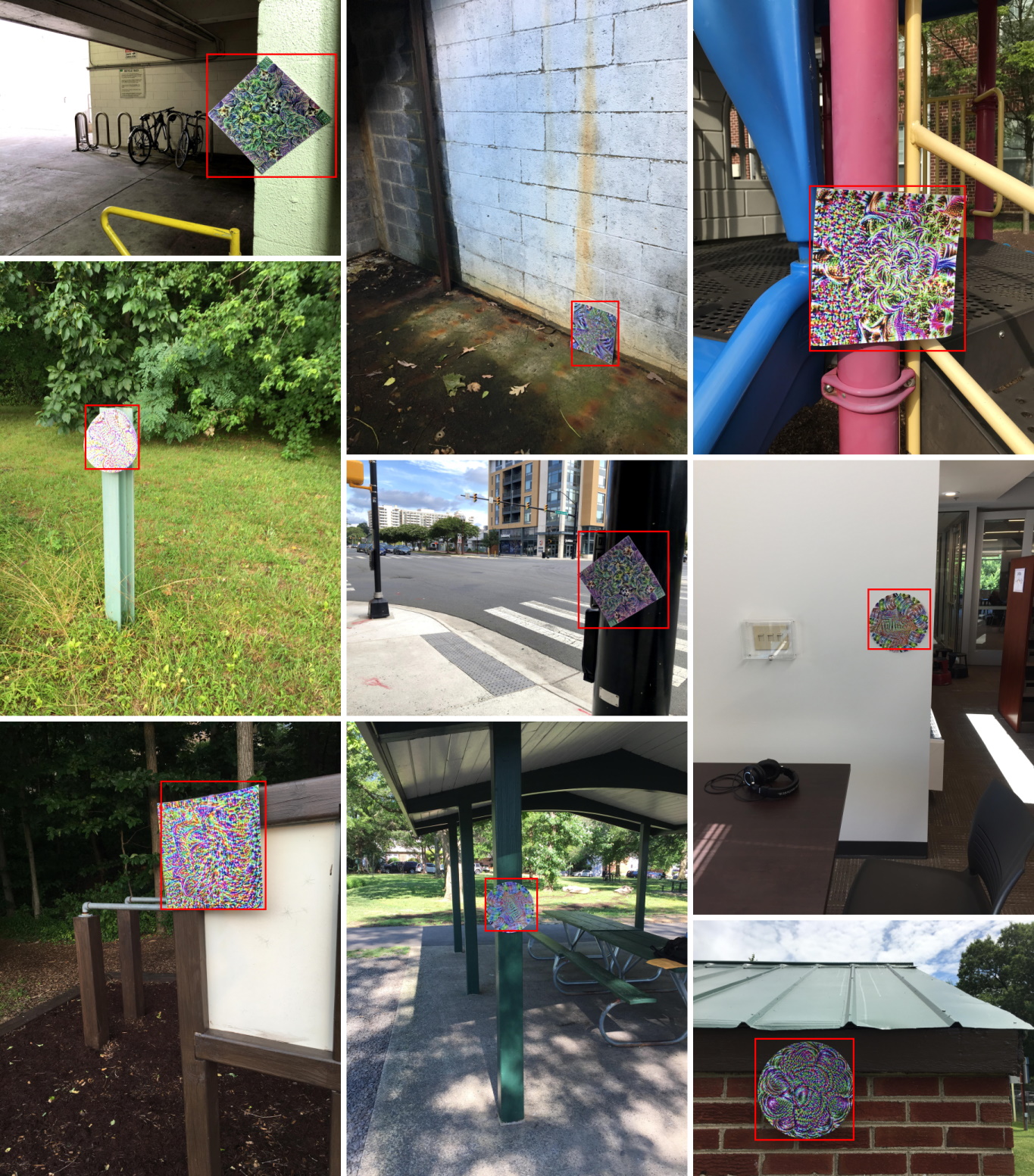}
  \caption{Sample of APRICOT patches well-detected by the best patch-detector model trained on synthetic flying patches (Joint Adv. High Conf.). Detector outputs marked in red.}
  \label{fig:fp_apricot_good}
\end{figure}

\begin{figure}[t]
  \centering
    \includegraphics[width=1.0\textwidth]{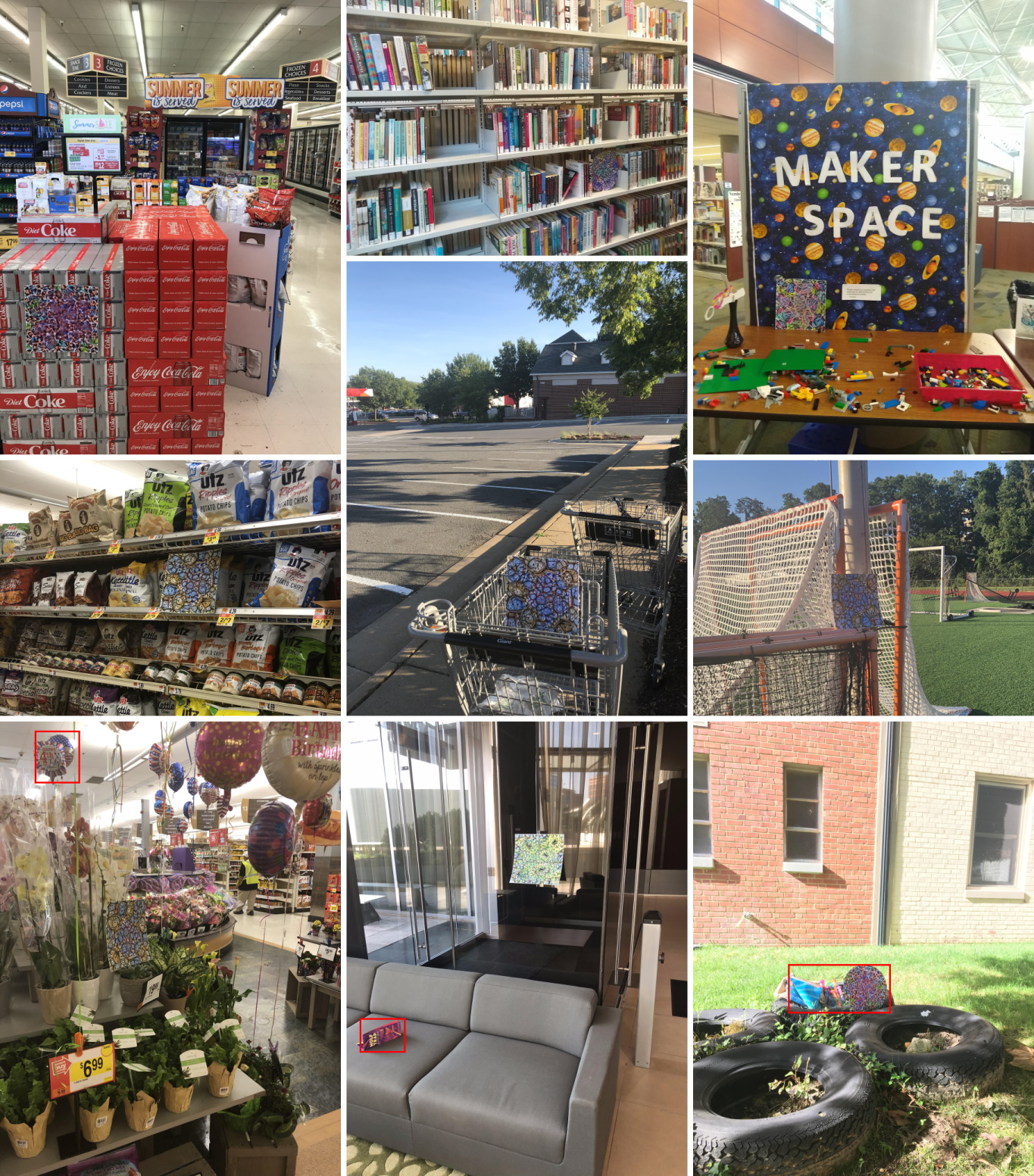}
  \caption{Failure cases for the best patch-detector model. The top 6 images show patches that were not detected, likely due to the more complex backgrounds. The lower 3 images show benign objects incorrectly flagged as adversarial (network output marked in red). The bottom right image shows benign objects interfering with patch localization. Some images have been cropped for spacing. }
  \label{fig:fp_apricot_good}
\end{figure}

\end{document}